\documentclass[runningheads]{llncs}

\usepackage[T1]{fontenc}
\usepackage{graphicx}
\usepackage{amssymb}
\usepackage{booktabs}
\usepackage{url}
\usepackage{hyperref}

\begin{document}

\title{Rashomon Memory:\\Towards Argumentation-Driven Retrieval for Multi-Perspective Agent Memory}
\titlerunning{Rashomon Memory}

\author{Albert Sadowski\orcidID{0009-0001-4524-5488}\thanks{Corresponding author.} \and \\
Jaros{\l}aw A. Chudziak\orcidID{0000-0003-4534-8652}}

\authorrunning{Sadowski and Chudziak}

\institute{Warsaw University of Technology, Poland\\
\email{\{albert.sadowski.stud,jaroslaw.chudziak\}@pw.edu.pl}}

\maketitle

\begin{abstract}
AI agents operating over extended time horizons accumulate experiences that serve multiple concurrent goals, and must often maintain conflicting interpretations of the same events. A concession during a client negotiation encodes as a ``trust-building investment'' for one strategic goal and a ``contractual liability'' for another. Current memory architectures assume a single correct encoding, or at best support multiple views over unified storage. We propose Rashomon Memory: an architecture where parallel goal-conditioned agents encode experiences according to their priorities and negotiate at query time through argumentation. Each perspective maintains its own ontology and knowledge graph. At retrieval, perspectives propose interpretations, critique each other's proposals using asymmetric domain knowledge, and Dung's argumentation semantics determines which proposals survive. The resulting attack graph is itself an explanation: it records which interpretation was selected, which alternatives were considered, and on what grounds they were rejected. We present a proof-of-concept showing that retrieval modes (selection, composition, conflict surfacing) emerge from attack graph topology, and that the conflict surfacing mode, where the system reports genuine disagreement rather than forcing resolution, lets decision-makers see the underlying interpretive conflict directly.

\keywords{Explainable AI \and Argumentation \and Multi-Agent Negotiation \and Conflict Resolution \and Agent Memory.}
\end{abstract}

\section{Introduction}
\label{section:introduction}

Consider an AI agent that supports a team preparing for a client negotiation. The team made a 15\% concession six months ago. Now they need to decide whether to extend similar terms to the client's subsidiary. What should the agent retrieve, and how should it explain its recommendation?

\begin{figure}[t]
    \centering
    \includegraphics[width=0.9\columnwidth]{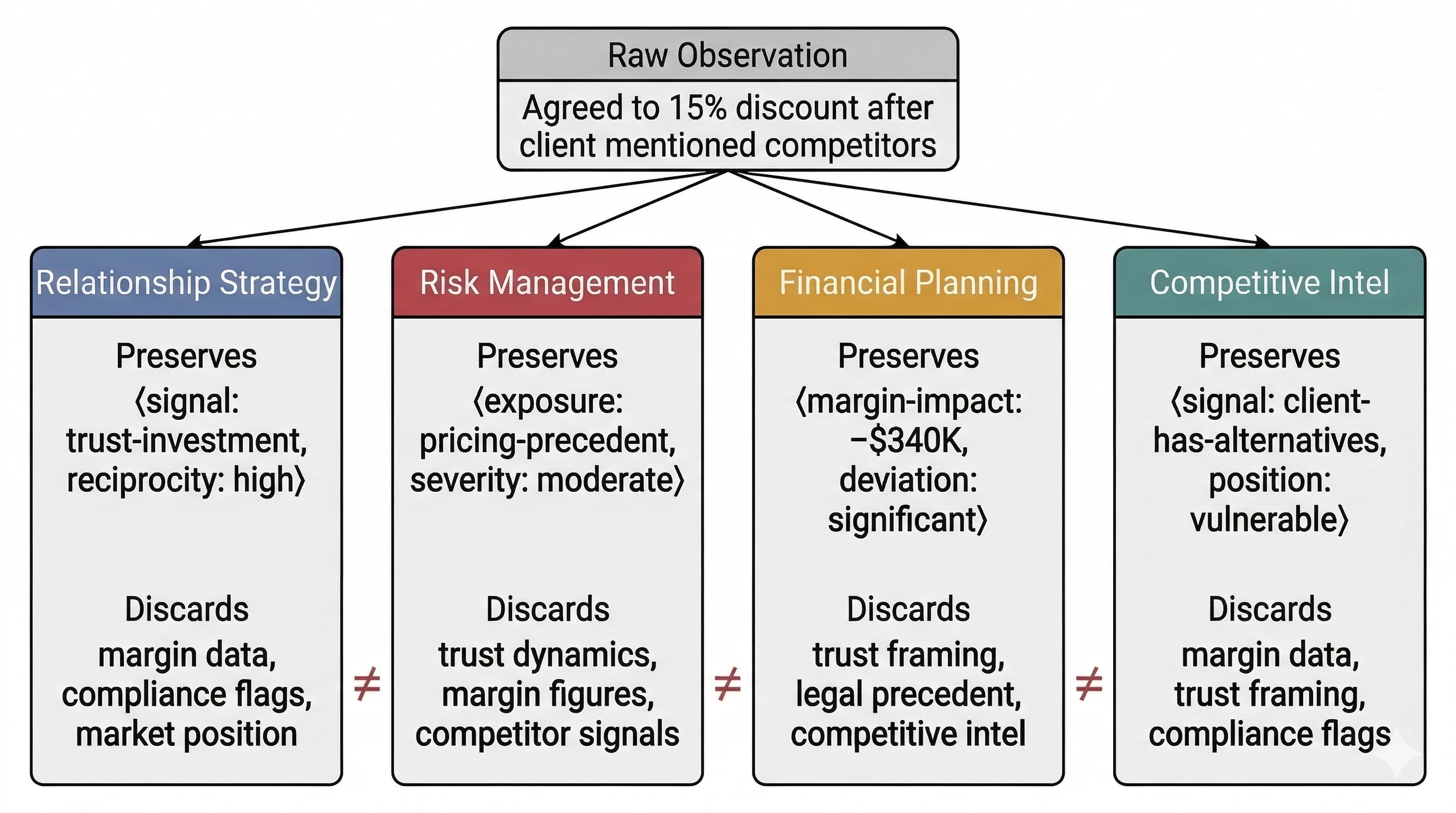}
    \caption{Parallel encodings of the concession event. Each perspective preserves what matters for its goals and discards what would interfere.}
    \label{fig:observation}
\end{figure}

The answer depends on which strategic concern is being served. From a relationship strategy perspective, the concession was a trust-building investment that created reciprocity expectations~\cite{lewicki2013trust,malhotra2004trust}. From a risk management perspective, it was a contractual liability that set a pricing precedent~\cite{worldcommerce2020}. Financial planning sees margin erosion. These are not different queries over the same record. They are conflicting interpretations that foreground different aspects of the same experience (Figure~\ref{fig:observation}). Any multi-goal agent must maintain these competing framings and, when queried, explain which framing it selected and why the alternatives were set aside. This is simultaneously a memory problem and an explainability problem.

Current memory architectures for LLM-based agents handle neither well. Systems that interpret at encoding time, knowledge graphs with fixed schemas, have to commit to one perspective before knowing which will matter~\cite{jiang2025EvolutionKnowledgeGraphsSurvey}. Systems that defer interpretation to retrieval, vector stores queried through retrieval-augmented generation, preserve optionality but never accumulate goal-relevant structure~\cite{lewis2020RAGForKnowledgeIntensiveNLPTasks,park2023GenerativeAgentsInteractiveSimulacraOfHumanBehavior}. The raw experience ``agreed to 15\% discount after client mentioned competitors'' doesn't use risk vocabulary. A query oriented toward risk exposure might not retrieve it at all, since embedding-based retrieval matches on semantic similarity, not goal-relevance~\cite{qian2024MemoRAG}. Recent work addresses retrieval limitations through dynamic forgetting~\cite{zhong2023memorybankenhancinglargelanguage}, self-critique~\cite{asai2023selfraglearningretrievegenerate}, and hierarchical summarization~\cite{sarthi2024raptorrecursiveabstractiveprocessing}, but these still assume retrieved content serves a single interpretive frame. None provide the user with an account of what alternative framings existed and why one was preferred.

There is cognitive science behind this. Retrieval success depends on alignment between encoding context and retrieval context~\cite{tulving1973encoding}. Morris et al.\ showed that encoding for one purpose produces memories poorly suited for another, even when the ``deeper'' encoding would typically be considered superior~\cite{morris1977levels}. Minsky argued that minds are better understood as societies of agents with competing priorities than as unified systems~\cite{minsky1986SocietyOfMind}. For agents that support negotiation, this means maintaining multiple strategic perspectives and surfacing the relevant one depending on context.

Our proposal: goals should parameterize encoding, not just retrieval. Each perspective preserves what matters for its goals and discards what would interfere. When \textit{Relationship Strategy} encodes the concession as a trust signal, it discards margin information. When \textit{Financial Planning} encodes it as margin erosion, it discards relationship framing. The resulting encodings are separate memories shaped by separate goals. Memory for multi-goal agents is therefore a coordination problem among encoding agents with conflicting priorities.

This paper proposes \emph{Rashomon Memory}\footnote{Named for Kurosawa's 1950 film \emph{Rashomon}}: an architecture where parallel goal-conditioned agents encode experiences according to their priorities and negotiate at query time through argumentation. The negotiation produces an attack graph that both determines which interpretation to surface and explains why that interpretation was chosen over alternatives~\cite{miller2019explanation}. When perspectives genuinely conflict, the system surfaces the disagreement rather than forcing resolution, letting the decision-maker see the competing framings directly.

Our contributions: (1) a problem formulation recasting AI agent memory as multi-agent coordination, (2) an architectural proposal combining goal-conditioned encoding with distributed peer critique for negotiated recall, (3) a proof-of-concept demonstrating that retrieval modes and contrastive explanations emerge from domain-grounded critique resolved through Dung's argumentation semantics, and (4) a research agenda addressing scalable argumentation, evolutionary ontology management, and user-facing explanation design.

\section{Rashomon Memory: Architecture and Components}

This section presents the Rashomon Memory architecture. The core idea is that each goal maintains its own encoding of shared experiences, and conflicting interpretations are resolved through argumentation at query time. We first describe the three components of the architecture, then explain how they interact during retrieval.

\begin{figure}[h!]
  \centering
  \includegraphics[width=\columnwidth]{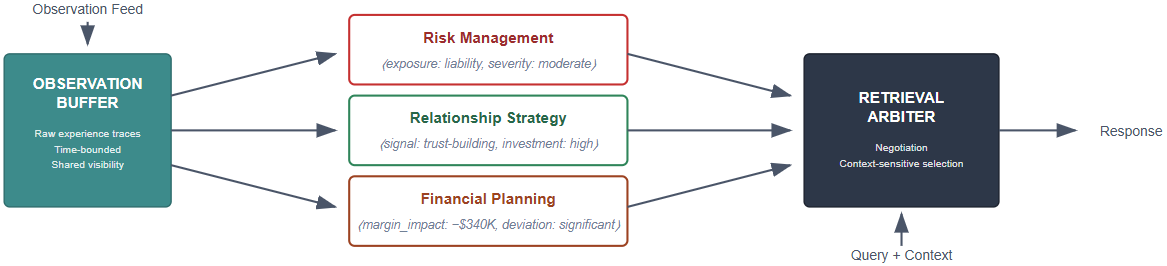}
  \caption{The Rashomon Memory architecture. Raw experiences enter a shared Observation Buffer, where parallel Goal-Perspectives encode them through distinct ontologies into separate knowledge graphs. The Retrieval Arbiter negotiates among perspectives at query time through argumentation, surfacing interpretations that fit the query context, along with an account of what was rejected and why.}
  \label{fig:diagram}
\end{figure}

Figure~\ref{fig:diagram} shows the proposed architecture. Three components interact: an \textit{Observation Buffer} where raw experiences await interpretation, \textit{Goal-Perspectives} that encode experiences according to their priorities, and a \textit{Retrieval Arbiter} that negotiates among perspectives at query time. We describe each below.

\subsection{Observation Buffer}

Raw experiences enter a time-bounded buffer visible to all goal-perspectives. This buffer isn't memory itself, it's a shared log from which perspectives construct their encodings.

Minsky's K-lines offer a useful analogy here: they connect agents that were active during an experience, enabling later reactivation~\cite{minsky1986SocietyOfMind}. Our buffer plays a similar staging role. It holds experiential traces long enough for multiple encoding processes to operate, then releases them. What persists isn't the buffer content but the perspective-specific encodings it enabled.

The buffer faces practical constraints. Experiences need to persist long enough for all relevant perspectives to process them, but not so long that the buffer becomes a de facto memory store. Biological memory systems exhibit similar staging: hippocampal structures hold experiences temporarily before cortical consolidation~\cite{schacter2001seven}. We leave optimal retention policies to future work.

\subsection{Goal-Perspectives}

Each goal-perspective maintains three components: encoding agents that transform buffer contents into structured representations, an ontology defining the vocabulary for encoding, and a knowledge graph accumulating encoded experiences. Figure~\ref{fig:observation} shows how four perspectives may encode the same observation.

\paragraph{Ontology Construction.}

Each perspective needs an ontology, a vocabulary of concepts and relations suited to its goals. Where do these ontologies come from? We see three possibilities, probably used in combination.

First, ontologies can be seeded by domain experts or inherited from existing taxonomies. \textit{Risk Management} might adopt categories from compliance frameworks; \textit{Financial Planning} might inherit from accounting standards. This gives you initial structure but can't anticipate every concept a perspective will need.

Second, ontologies can emerge from encoding patterns. When encoding agents repeatedly struggle to categorize an experience, that signals a conceptual gap. Ontology learning from text is an active research subject~\cite{wong2012ontology}, and recent work uses LLMs to extract taxonomic relations from corpora~\cite{babaei2023llmontology}. A perspective could apply similar techniques to its own encoding history, identifying clusters that lack named concepts; recent work confirms that LLM-driven ontology construction can yield representations sufficient for downstream symbolic reasoning~\cite{sadowskiChudziakOnVerifiable}.

Third, ontologies can evolve through use. Jiang et al.\ survey how knowledge graphs have evolved from static structures to temporal and event-centric representations~\cite{jiang2025EvolutionKnowledgeGraphsSurvey}. A key finding: schema rigidity creates brittleness. Graphs that can't extend their ontologies degrade as domains shift. Our architecture assumes ontologies are living structures, refined asynchronously as new experiences reveal gaps or redundancies.

\paragraph{Knowledge Graph Curation.}

Each perspective maintains a separate knowledge graph populated through its encoding process. This raises practical questions about storage and maintenance.

For persistence, existing temporal knowledge graph infrastructure offers a starting point. StreamE supports continuous updates to graph structure as new facts arrive~\cite{zhu2023streamE}. Zep maintains temporal indices over knowledge graphs for agent memory~\cite{rasmussen2025zep}. These systems assume unified storage, but their update mechanisms could serve perspective-specific graphs.

Curation involves more than storage, though. As encodings accumulate, perspectives have to manage redundancy, resolve internal contradictions, and forget outdated information. We propose that retrieval acts as a curation signal: encodings that get frequently retrieved gain importance weight; those never retrieved decay. This creates a stigmergic dynamic where use patterns shape memory structure, similar to how ant pheromone trails reinforce successful paths. Kirkpatrick et al.\ demonstrated importance-weighted preservation in neural networks through elastic weight consolidation~\cite{kirkpatrick2017EWC}; an analogous mechanism could protect frequently-accessed encodings from decay while letting unused ones fade.

\paragraph{Perspective Boundaries.}

One design question we leave open: how many perspectives, and who decides their boundaries? The examples in this paper assume perspectives aligned with organizational functions (relationship management, risk, finance). But perspectives could be more fine-grained (specific client relationships) or more abstract (short-term vs.\ long-term optimization). Li et al.\ found that explicit belief states in multi-agent systems reduce hallucination by helping agents maintain distinct viewpoints~\cite{li2023theoryofmind,kostkaChudziakToM}, which suggests that perspective boundaries should be explicit and maintained, not emergent and fluid. We suspect the right granularity depends on the agent's operational context, but establishing principles for perspective design remains future work.

\subsection{Retrieval Arbiter}
\label{sec:retrievalArbiter}

Retrieval in this architecture is negotiation, not search. When a query arrives, the system doesn't traverse a unified index. Instead, it kicks off a structured protocol where perspectives propose interpretations, critique competing proposals, and an arbiter resolves which proposals survive. The architecture contributes two separable ideas: a \emph{distributed peer critique} pattern, in which domain-specialized agents evaluate each other's proposals using asymmetric knowledge, and a \emph{principled resolution mechanism} that computes acceptable sets from the resulting attack structure.

We fix terminology before describing the protocol. An \emph{observation} is a natural-language string describing an event; we use \emph{observation} and \emph{experience} as synonyms. A \emph{perspective} is a goal-conditioned LLM instance with a private ontology and knowledge graph. A \emph{query} is a natural-language input. A \emph{proposal} is an interpretation produced by a perspective from its own encodings, accompanied by a relevance claim; proposals are the arguments of the underlying argumentation framework. An \emph{attack} from proposal $a$ against proposal $b$ is a claim that $b$'s framing would mislead the querier in the current query context. The \emph{attack graph} is the pair $(A, R)$ where $A$ is the set of proposals submitted for the query and $R \subseteq A \times A$ is the set of attacks. The arbiter computes Dung's grounded extension over $(A, R)$. Proposals and attacks are LLM-generated, so $(A, R)$ is a heuristic construction; the resolution over a fixed $(A, R)$ is deterministic.

The protocol has four phases. First, the arbiter broadcasts the query, along with context information (querier identity, decision type, current priorities), to all perspective agents. Second, perspectives with relevant encodings submit proposals, candidate interpretations accompanied by relevance claims. Third, proposals get shared among perspectives, and each may submit attacks on others' proposals. An attack is a claim that, given the query context, the attacking perspective's framing should dominate. Fourth, the arbiter constructs an attack graph from the submitted attacks and computes an acceptable extension using Dung's argumentation semantics~\cite{dung1995acceptability}.

Each perspective knows its own domain and can evaluate competing proposals from a position of genuine expertise. When \textit{Risk Management} attacks \textit{Relationship Strategy's} proposal, it does so with access to compliance encodings that \textit{Relationship Strategy} never stored. This asymmetry is a feature: the attacking perspective has domain-specific knowledge that the defending perspective lacks, making the attack grounded rather than generic. The arbiter, by contrast, performs no reasoning about the content of the attacks. It applies Dung's grounded semantics to the attack graph as a lightweight, deterministic resolution step. This is a deliberate division of labor: the LLM constructs the attack graph, and the formal semantics resolves it~\cite{baroni2018handbook}. The construction is heuristic and prompt-dependent; the resolution is principled and deterministic. The formal layer earns its place by giving uniqueness of the grounded extension for any fixed graph~\cite{dung1995acceptability}, determinism of the retrieval mode given the constructed graph, and an attack graph that doubles as an explanation by construction. At three perspectives the topology is small enough that simple heuristics would often agree; at larger $N$ the topology is no longer trivial and a principled resolution is needed. Reliability of the constructed attack graph is what the architecture inherits from its LLM substrate, and is the central empirical open problem (Section~\ref{section:demonstration}).

This protocol relates to multi-agent debate~\cite{du2023MultiAgentDebate,liang2024MAD,irving2018debate,baba2026} but differs in its decision procedure. Debate frameworks typically use iterative refinement, where agents revise positions across rounds, converging toward agreement. Our protocol uses a single round of critique followed by extension computation over a static attack graph. In principle, debate could be configured to select among alternatives rather than converge toward a single answer. The structural difference is that our approach separates the generation of critiques (by domain-expert perspectives) from the resolution of those critiques (by a formal semantics), rather than interleaving them.

The attack graph that results from this protocol is both a resolution mechanism and an explanation artifact. Miller argues that human-interpretable explanations are contrastive: people ask ``why P rather than Q?'' rather than ``why P?''~\cite{miller2019explanation}. The attack graph answers exactly this kind of question. When \textit{Risk Management} is selected over \textit{Relationship Strategy}, the user can inspect the attack: \textit{Risk Management} argued that relationship framing omits compliance exposure, and this attack was not counterattacked. The explanation is grounded in the perspectives' own domain knowledge rather than generated post-hoc. This principle, that explanation can emerge from inspectable intermediate structure rather than requiring a separate generation step, has been validated in neural-symbolic settings, where externalising every interpretive decision as a formal assertion yields auditable reasoning traces by construction~\cite{sadowskiChudziak}. {\v{C}}yras et al.\ survey how argumentation frameworks naturally support this kind of reasoning trace~\cite{cyras2021argumentation}. In our architecture, the attack graph plays an analogous role: the grounded extension identifies which arguments survived and which were defeated, giving the user a justification structure rather than just an answer. Richer structured argumentation frameworks such as ASPIC+~\cite{modgil2014aspic} could further enrich the attack relation with explicit premises and inference rules. We leave this extension to future work and use abstract argumentation in the current prototype.

The quality of the architecture hinges on the quality of the peer critique. Each perspective must assess not just its own relevance but why competing framings are less appropriate for the current context. This is a form of theory of mind: perspectives benefit from models of each other's goals and typical proposal patterns. Li et al.\ found that explicit belief state representations improve coordination in multi-agent systems by helping agents reason about others' viewpoints~\cite{li2023theoryofmind}. In our architecture, such representations serve not cooperation but well-grounded critique. The advantage is that each perspective evaluates others from its own accumulated knowledge, not from a shared or neutral vantage point. The setup resembles argumentation-based negotiation, where agents justify positions through reasons grounded in their own knowledge bases~\cite{rahwan2005argumentation}. But our perspectives are not negotiating with each other for mutual benefit. They are competing to frame the answer to a query, and the user benefits from the resulting explanation.

\paragraph{Example: Query Walkthrough.}
Say a query arrives six months after the concession: ``Should we extend similar terms to Meridian's subsidiary?'' Each perspective searches its encodings and proposes interpretations. \textit{Relationship Strategy} retrieves the concession as a trust-building gesture and proposes extending terms to reinforce the partnership. \textit{Financial Planning} retrieves the same event as margin erosion and proposes caution about compounding losses. \textit{Risk Management} retrieves it as precedent-setting and flags legal exposure from inconsistent treatment of related entities. These proposals enter the peer critique phase. \textit{Financial Planning} attacks \textit{Relationship Strategy}: the query concerns terms, which implies cost analysis; relationship framing omits margin impact. \textit{Relationship Strategy} counters: the phrase ``similar terms'' presupposes the original concession as reference; cost analysis alone ignores why the precedent exists. The arbiter then computes which proposals survive given the resulting attack graph. If only \textit{Financial Planning} survives, it dominates. If multiple survive, the system composes them. If all attack each other, it surfaces all three framings rather than forcing resolution. The same event surfaces through three different framings. Which one gets selected depends on why the question is being asked.

\subsection{Retrieval Modes}

Not all queries resolve to a single frame. Take ``Which clients show early signs of churn?'' No single perspective owns churn prediction. \textit{Relationship Strategy} contributes signals about reduced executive engagement. \textit{Financial Planning} contributes pricing sensitivity patterns. \textit{Risk Management} contributes audit demand frequency. \textit{Competitive Intelligence} contributes mentions of competitor conversations. The arbiter's role here is synthesis: assembling complementary signals that no perspective could provide alone. Minsky's ``non-compromise principle'' guides this synthesis~\cite{minsky1986SocietyOfMind}. Rather than averaging perspectives into undifferentiated mush, the system lets each contribute what it uniquely preserved. Composition isn't consensus; it's structured aggregation of distinct viewpoints.

Other queries produce genuine conflict that resists both selection and composition. ``Was the Meridian renewal a success?'' elicits four incompatible answers: \textit{Relationship Strategy} says yes (trust deepened), \textit{Financial Planning} says no (margin eroded), \textit{Risk Management} says ambiguous (exposure contained but precedent set), \textit{Competitive Intelligence} says qualified yes (retained against threat). The arbiter must not synthesize false consensus here. Some queries simply have no perspective-independent answer~\cite{haraway1988situated,nagel1986view}. The appropriate response is to surface the conflict itself, making the disagreement legible rather than hidden.

This is the core insight behind Rashomon Memory, named after Kurosawa's film in which the same event receives irreconcilable accounts from different witnesses~\cite{Richie1987Rashomon}. The architecture supports three retrieval modes: \emph{selection} when context determines a dominant frame, \emph{composition} when perspectives contribute complementary information, and \emph{surfacing} when genuine conflict should be reported rather than resolved. These modes are not designed directly; they emerge from the topology of the attack graph produced by Dung's grounded semantics. This is where argumentation earns its role in the architecture. The attack graph both \emph{resolves} which interpretation to surface and \emph{explains} why: each attack articulates a domain-grounded reason for preferring one framing over another, and the grounded extension records which reasons survived. The surfacing mode makes this most visible. Most decision support systems converge on a single recommendation, leaving the user no account of what alternatives existed. When perspectives genuinely conflict, the attack graph preserves the disagreement as a structured, inspectable object and the user sees not just that the system could not decide, but \emph{on what grounds} each perspective challenged the others. The argumentation trace is the explanation.

Separate families of related work clarify what is and isn't novel here. Retrieval-augmented generation~\cite{lewis2020RAGForKnowledgeIntensiveNLPTasks} retrieves over a single semantic frame and does not condition encoding on goals; goal-relevance has to be extracted at query time from a representation that did not preserve it. Knowledge graphs with fixed schemas~\cite{jiang2025EvolutionKnowledgeGraphsSurvey} commit to a single ontology before observing which goal will matter, and cannot maintain conflicting interpretations of the same event. Multi-agent debate~\cite{du2023MultiAgentDebate,liang2024MAD} couples critique and convergence in the same iterative loop, aiming at a single agreed answer; we separate critique from resolution and let a formal semantics over a static attack graph carry the resolution, which makes the explanation an inspectable artifact rather than an iterative trace and lets the system surface conflict as a first-class outcome rather than forcing convergence. In the following section, we demonstrate this on a controlled scenario.

\section{Case Study: Implementation and Demonstration}
\label{section:demonstration}

We implemented a prototype of Rashomon Memory to test whether the architecture produces sensible behavior on a controlled scenario. The encoding agents use the LLMs4OL methodology~\cite{babaei2023llmontology} for ontology learning, with each perspective maintaining its own OWL/RDF knowledge graph in Turtle format. The retrieval arbiter uses Dung's abstract argumentation semantics~\cite{dung1995acceptability}. All LLM calls use \href{https://platform.claude.com/docs/en/about-claude/models/overview}{Claude Sonnet 4.6} accessed via Anthropic API; model hyperparameters set to defaults, except for the temperature - set to 0. The scenario extends the negotiation example from Section~\ref{section:introduction}: three perspectives (\textit{Relationship Strategy}, \textit{Risk Management}, \textit{Financial Planning}) encode eight observations from a client engagement with Meridian Corp, after which we run four queries against the accumulated knowledge graphs. The source code is publicly available.\footnote{\url{https://github.com/albsadowski/rashomon-memory-demo}}

\paragraph{Goal-Conditioned Encoding.}

Each perspective receives every observation but independently decides whether to encode it. The pipeline has two stages. A relevance filter determines whether the observation matters for this perspective's goals. For observations that pass, the encoding agent follows the LLMs4OL methodology: term typing (Task~A) classifies entities against the perspective's TBox, taxonomy discovery (Task~B) places any new classes in the hierarchy, and relation extraction (Task~C) identifies properties between entities. The output is a set of RDF triples added to the perspective's ABox, along with any TBox extensions needed to express them. Each perspective is implemented as a separate LLM call with a goal-specific system prompt and a structured output schema that returns both ontology updates and instance triples.

\begin{table}[t]
\small
\centering
\caption{Encoding selectivity across observations and perspectives. Of 24 possible encodings, 13 passed the relevance filter. Each perspective encodes a different subset, and perspectives that encode the same observation extract different information.}
\begin{tabular}{p{8cm}ccc}
\toprule
\textbf{Observation} & \textbf{Rel.} & \textbf{Risk} & \textbf{Fin.} \\
\midrule
1. Competitor proposals at quarterly review & $\checkmark$ & -- & -- \\
2. 15\% discount concession granted & $\checkmark$ & $\checkmark$ & $\checkmark$ \\
3. VP praises support team response & $\checkmark$ & -- & -- \\
4. Audit flags discount exceeding policy band & -- & $\checkmark$ & $\checkmark$ \\
5. Account margin drops 34\% $\to$ 22\% & -- & -- & $\checkmark$ \\
6. Subsidiary inquires about platform adoption & $\checkmark$ & -- & $\checkmark$ \\
7. CEO praises partnership at industry summit & $\checkmark$ & -- & -- \\
8. Legal notes precedent risk across Tier-2 clients & -- & $\checkmark$ & $\checkmark$ \\
\midrule
\textbf{Encoded / Total} & 5/8 & 3/8 & 5/8 \\
\bottomrule
\end{tabular}
\label{tab:selectivity}
\end{table}

Table~\ref{tab:selectivity} shows the encoding decisions. The pattern reflects perspective goals. \textit{Relationship Strategy} encoded stakeholder interactions and partnership signals (observations 1, 2, 3, 6, 7) but skipped audit findings, margin reports, and legal reviews. \textit{Risk Management} encoded precedent-setting events and compliance flags (observations 2, 4, 8). \textit{Financial Planning} tracked margin and revenue impacts (observations 2, 4, 5, 6, 8). The selectivity extends beyond binary filtering: perspectives that encode the same observation extract different information. Consider observation~2, the 15\% discount concession. \textit{Relationship Strategy} encoded it as a trust signal with ``guarded'' trust level and ``low'' reciprocity expectation, noting the concession was extracted through competitor leverage rather than offered as goodwill. \textit{Risk Management} encoded it as a pricing precedent with ``moderate'' severity, flagging policy deviation. \textit{Financial Planning} encoded it as a margin impact of $-15\%$. Each perspective discards what the others preserve. The resulting knowledge graphs are not views over shared storage. They are separate memories with separate vocabularies, shaped by separate goals.

\paragraph{Retrieval Through Argumentation.}

The retrieval pipeline implements the protocol described in Section~\ref{sec:retrievalArbiter}, combining distributed peer critique with Dung's grounded semantics for resolution. Each perspective searches its knowledge graph and submits a proposal: an interpretation of its relevant encodings, accompanied by a relevance claim. Perspectives then evaluate each other's proposals and may submit attacks. An attack is a claim that another perspective's framing would mislead the querier given the specific query context. The attacks form a directed graph, and Dung's grounded semantics determines which proposals survive. Both proposal and attack calls run at temperature 0 with strict JSON output schemas: a proposal returns a natural-language interpretation paired with a relevance-claim string; an attack returns an attacker-target-rationale triple. The attack graph $(A, R)$ is built from these triples and passed to the grounded-extension algorithm. We use LLMs at every stage of the pipeline (relevance filtering, encoding, proposal, attack) because all four require open-domain natural-language reasoning over heterogeneous inputs, for which no mature symbolic alternative exists.

The peer critique step is where the inferential work happens. Each perspective evaluates others from its own domain knowledge. When \textit{Risk Management} attacks \textit{Relationship Strategy}, it draws on compliance and precedent encodings that \textit{Relationship Strategy} never stored. The attack is grounded in asymmetric expertise, not generic disagreement. Two properties are needed for this to work. First, attacks must be selective: a perspective that attacks everything degrades the system into trivial mutual conflict. Second, attacks must be contextual: the same two perspectives might or might not attack each other depending on the query. Both properties are encouraged through prompt design: perspectives are instructed to attack only when they have a genuine, query-grounded reason, and to leave complementary proposals unattacked. Whether the LLM reliably follows these instructions is an empirical question. Multi-agent debate systems without such constraints tend to suffer from semantic drift and logical deterioration~\cite{maslowskiChudziak}, which is what the selectivity and contextuality properties are designed to prevent. The results are encouraging: the same three perspectives produced attack graphs with zero, two, three, and six edges across four queries (Table~\ref{tab:query_results}), with no manual tuning of the attacks.

The resolution step is deliberately lightweight. Dung's grounded extension is computed iteratively: begin with all unattacked arguments, then add any argument whose attackers are all counterattacked by arguments already in the set, and repeat until stable. This process converges to a unique fixed point, making the retrieval mode deterministic for a given attack graph. If the grounded extension contains all proposals, no perspective was successfully challenged, and the system composes their contributions (\emph{composition}). If it contains a single proposal, that perspective dominates (\emph{selection}). If it is empty, every perspective is caught in unresolved mutual conflict, and the system surfaces the disagreement rather than forcing resolution (\emph{surfacing}). Multiple survivors that do not exhaust all proposals represent a filtered \emph{composition}. At the scale of our demonstration (three perspectives), the same outcomes could be reached by simpler heuristics. The grounded extension is unique for any attack graph, which makes the retrieval mode deterministic given the perspectives' critiques. This separates the concern of generating critiques from the concern of resolving them. We illustrate the mechanism with two queries that produce contrasting outcomes.

\begin{figure}[t]
    \centering
    \includegraphics[width=0.9\columnwidth]{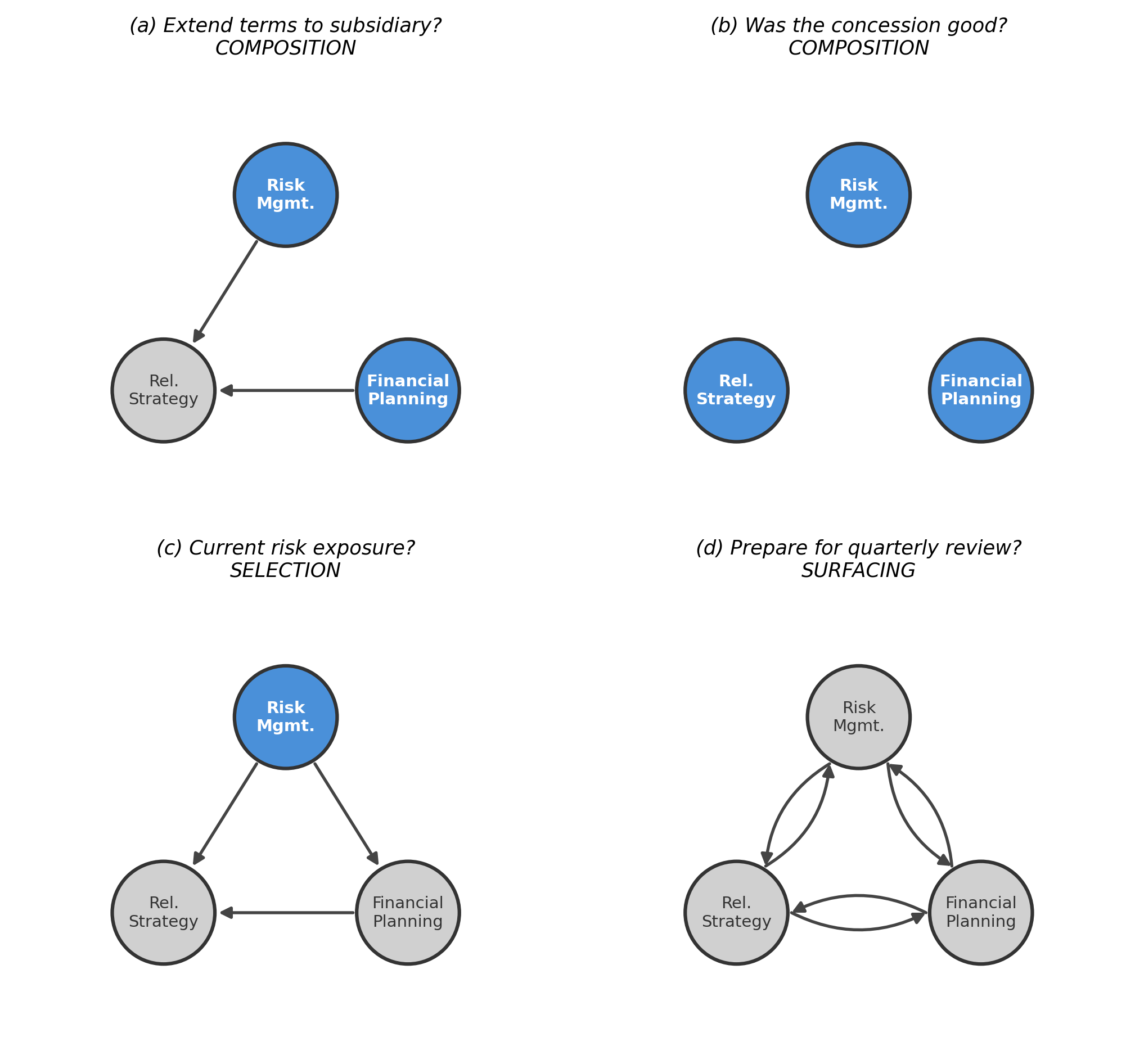}
    \caption{Attack graphs for four queries. Arrows denote attacks. Following Caminada's labelling under grounded semantics~\cite{caminada2006reinstatement}, blue nodes are \emph{in} (accepted); grey nodes are \emph{out} (defeated by an accepted argument) in (a) and (c), and \emph{undecided} in (d), where the grounded extension is empty. (a)~Risk and Financial attack Relationship; survivors compose. (b)~No attacks; all compose. (c)~Risk Management attacks both and is sole survivor (selection). (d)~Complete mutual attacks; empty grounded extension, all nodes undecided (surfacing).}
    \label{fig:attack_graphs}
\end{figure}

\textit{``What is our current risk exposure from the Meridian account?''} All three perspectives proposed interpretations. \textit{Risk Management} attacked \textit{Relationship Strategy}: ``Relationship Strategy frames exposure through trust levels and reciprocity dynamics. These are valid relationship metrics, but they do not constitute risk exposure in the compliance, legal, or financial sense the query requires.'' \textit{Risk Management} also attacked \textit{Financial Planning}: ``Financial Planning quantifies margin erosion, but margin impact is a financial performance metric, not a risk exposure assessment.'' \textit{Financial Planning} independently attacked \textit{Relationship Strategy} on similar grounds. \textit{Relationship Strategy} attacked neither. The resulting graph (Figure~\ref{fig:attack_graphs}c) has \textit{Risk Management} as the only unattacked node. The grounded extension computation confirms this: \textit{Risk Management} enters first (unattacked, therefore trivially acceptable). \textit{Financial Planning} cannot enter because its attacker (\textit{Risk Management}) is not counterattacked by anyone already in the extension. \textit{Relationship Strategy} is excluded for the same reason. The grounded extension is \{\textit{Risk Management}\}, and the mode is \emph{selection}.

\textit{``How should we prepare for the upcoming Meridian quarterly review?''} Again all three perspectives proposed, but the attack structure looked very different. \textit{Relationship Strategy} argued the review should focus on partnership repair and avoid anchoring on the discount. \textit{Risk Management} argued compliance remediation should take priority. \textit{Financial Planning} argued margin recovery should dominate. Each perspective attacked both others. \textit{Relationship Strategy} attacked \textit{Risk Management}: ``treating the review as a compliance exercise risks alienating the client.'' \textit{Risk Management} attacked \textit{Relationship Strategy}: ``advising to avoid discussing the discount leaves compliance violations unaddressed.'' All six possible directed edges appeared (Figure~\ref{fig:attack_graphs}d). The grounded extension is empty: no perspective is unattacked, so the characteristic function admits nothing. Each perspective is nonetheless self-defending, it attacks both of its attackers, so the preferred extensions are the three singletons, each containing exactly one perspective. Under Caminada's labelling all three arguments are undecided rather than out, which is the precise condition that distinguishes surfacing from selection~\cite{caminada2006reinstatement}. The mode is \emph{surfacing}. Rather than forcing a single response, the system reports the disagreement itself: all three framings are presented as legitimate yet mutually attacking, and the attack graph records the grounds on which each perspective challenged the others. How to prepare for the review genuinely depends on what the team is optimizing for.

The remaining two queries complete the picture. ``Should we extend similar discount terms to the subsidiary?'' produced two attacks against \textit{Relationship Strategy} (from \textit{Risk Management} and \textit{Financial Planning}), yielding a grounded extension of \{\textit{Risk Management}, \textit{Financial Planning}\} and \emph{composition} of the two surviving perspectives (Figure~\ref{fig:attack_graphs}a). ``Was the concession a good decision?'' produced no attacks. All three perspectives contributed evaluative assessments without finding the others' framings misleading for this query context. The grounded extension contained all three proposals, and the mode was again \emph{composition}, though for a different reason: complementary rather than pruned (Figure~\ref{fig:attack_graphs}b).

\begin{table}[t]
\small
\centering
\caption{Query results. The retrieval mode emerges from the grounded extension of the attack graph constructed by the perspectives.}
\begin{tabular}{p{5cm}rll}
\toprule
\textbf{Query} & \textbf{Attacks} & \textbf{Grounded ext.} & \textbf{Mode} \\
\midrule
Extend terms to subsidiary? & 2 & \{Risk, Fin.\} & Composition \\
Was the concession good? & 0 & \{Rel., Risk, Fin.\} & Composition \\
Current risk exposure? & 3 & \{Risk\} & Selection \\
Prepare for quarterly review? & 6 & $\emptyset$ & Surfacing \\
\bottomrule
\end{tabular}
\label{tab:query_results}
\end{table}

\paragraph{User-Facing Explanation.}

What does the user actually receive? Consider the risk exposure query (selection mode). The system presents Risk Management's interpretation as the primary response, but accompanies it with an explanation derived from the attack graph: ``This response is based on Risk Management's assessment. Financial Planning's margin analysis and Relationship Strategy's trust framing were also considered but set aside. Risk Management argued that margin impact is a financial performance metric rather than a risk exposure assessment, and that trust levels do not constitute risk exposure in the compliance sense this query requires. Neither perspective counterattacked.'' In the surfacing case (quarterly review preparation), the system presents all three framings as alternatives: ``Three strategic perspectives apply and they conflict. Relationship Strategy recommends focusing on partnership repair. Risk Management recommends compliance remediation. Financial Planning recommends margin recovery. Each perspective argued that the others' framing would misdirect the review. The system cannot recommend one framing over the others without knowing which strategic priority the team is currently optimizing for.'' The attack graph (Figure~\ref{fig:attack_graphs}) is available for inspection in both cases. We have not conducted user studies to evaluate whether this form of explanation improves decision quality. Designing and evaluating such studies is a direction for future work.

\paragraph{Observations and Limitations.}

The central observation is that attack graph topology is query-dependent, not perspective-dependent. The same three perspectives produced four distinct topologies across four queries. This means the retrieval mode is determined by the interaction between query context and perspective goals, not by static properties of the perspectives themselves.

The encoding pipeline required 24 relevance checks and 13 encoding calls (37 LLM invocations total for 8 observations). Each query required up to 7 invocations (3 proposals, 3 attack evaluations, 1 response assembly). At current LLM latencies, this is feasible for deliberative tasks but not for interactive use.

We make no efficacy claim. This work demonstrates feasibility of the architecture, not improvement over baselines. We report no quantitative metrics, no comparison with simpler alternatives such as RAG over raw observations or single-perspective retrieval, no user studies of explanation quality, and no evaluation at larger numbers of perspectives or more diverse queries. The central reliability concern the architecture inherits is the LLM's behavior in attack generation: in our four queries we observed no indiscriminate attacking, but this property is not guaranteed by the architecture, and is the open empirical question on which a more substantial evaluation should focus.

\section{Discussion and Research Agenda}

The architecture has clear boundaries. Factual recall, e.g., \textit{what time is the meeting?}, \textit{what was the client's name?}, does not benefit from multiple interpretations. Domains with ground truth or regulatory compliance requirements may need unified encodings instead. The computational cost of argumentation at query time may be too high for latency-sensitive applications. The approach is best suited for deliberative settings, such as negotiation preparation, strategic planning, and post-mortem analysis, where the cost of a richer explanation is justified by the stakes of the decision. Within its scope, multi-perspective memory requires progress on several fronts.

\paragraph{Evolutionary Ontology.}

When goals shift, past encodings become obstacles. Consider a firm that spent two years encoding experiences through a growth-oriented ontology. \textit{Relationship Strategy} used concepts like ``trust-building'' and ``relationship investment''. Now the firm shifts to profitability focus. The old encodings still exist, but the vocabulary no longer fits current priorities. What does ``trust-building'' mean when every interaction must justify its margin impact? The ontology needs new concepts, or old encodings need reinterpretation through a lens that did not exist when they were created. The open problem is also retrospective reinterpretation. Can a perspective re-encode past experiences under an updated ontology? If so, what prevents perspectives from fabricating convenient histories?

\paragraph{Scalable Arbitration.}

The cost analysis below is analytical, not measured. The LLM invocation cost of the retrieval pipeline is dominated by the peer critique phase. With $N$ perspectives, each query requires up to $N$ proposal calls, $N$ attack evaluation calls (each perspective evaluates all others in a single call), and one response assembly call, giving $O(N)$ LLM invocations per query. If attack evaluation is pairwise (each perspective evaluates each other perspective in a separate call), the cost rises to $O(N^2)$. With 10 perspectives under pairwise evaluation, a single query would require 10 proposals + 90 attack evaluations + 1 assembly = 101 calls. Encoding scales as $O(N \cdot M)$ for $M$ observations, though the relevance filter reduces the constant. The extension computation itself is not the bottleneck: computing the grounded extension is polynomial in the number of arguments~\cite{dung1995acceptability}, and preferred extensions, while exponential in the worst case, operate over the number of perspectives (typically small), not the number of stored experiences.

Several mitigations are available. Batched evaluation (one call per perspective, evaluating all other proposals at once) keeps the query cost at $O(N)$ rather than $O(N^2)$. Selective participation, where perspectives abstain when they lack relevant encodings, reduces $N$ to the number of engaged perspectives per query. In our demonstration, not every perspective proposed for every query. More aggressive strategies could cluster perspectives hierarchically, running arbitration within clusters before a cross-cluster round. Whether these mitigations preserve attack quality at scale is an open question. Miscalibrated abstention could cause relevant perspectives to stay silent, while overconfident ones dominate.

\paragraph{Explanation Design.}

The attack graph provides the raw material for explanation, but how to present it to different users remains open. A negotiation team preparing for a meeting may want a brief summary of which perspective dominated and why. A compliance officer reviewing past decisions may want the full attack trace. Miller's work on contrastive explanation~\cite{miller2019explanation} suggests that the most useful form is often the minimal contrast: why was this framing selected rather than that one? The attack graph supports this directly, since each attack articulates a reason for preferring one framing over another. However, our current prototype presents explanations as unstructured text. Richer interfaces could visualize the attack graph itself, allow users to drill into specific attacks, or let users override the resolution by manually accepting a defeated perspective. Whether users find argumentation-based explanations more trustworthy or actionable than alternatives (e.g., confidence scores, feature attribution) is an empirical question that requires user studies. The architecture also treats retrieval as read-only. A future extension might let retrieval outcomes feed back into storage, reinforcing encodings that prove useful~\cite{lee2009reconsolidation}. Whether such path dependence risks collapsing plurality back into single-perspective dominance is a question we leave open.

\section{Conclusion}

This paper proposed Rashomon Memory, an architecture where parallel goal-conditioned agents encode experiences according to their priorities and negotiate at query time through argumentation. The proof-of-concept showed that attack graph topology is query-dependent. The same perspectives produced different retrieval modes across different queries, determined by the interaction between query context and perspective goals. The retrieval mode (selection, composition, or conflict surfacing) emerged from the interaction between query context and perspective goals, not from static properties of the perspectives themselves.

The conflict surfacing mode is the most distinctive outcome. When perspectives genuinely disagree, the system reports the disagreement as a structured, inspectable object rather than forcing a single answer. The decision-maker sees not only which interpretation was selected, but which alternatives were considered and on what grounds they were rejected.

Two contributions are separable from the architecture. A distributed peer critique pattern where domain-specialized agents evaluate each other's proposals using asymmetric knowledge. And Dung's grounded semantics as a resolution mechanism that doubles as a justification structure. Together they connect retrieval to explanation without a separate generation step.

The mechanisms for ontology evolution, scalable arbitration, and user-facing explanation design remain open. We have not compared the architecture against baseline retrieval methods on standard metrics. These are directions for future work that follow from the problem formulation.

\bibliographystyle{splncs04}
\bibliography{references}

\end{document}